\documentstyle[aclap]{article}

\title{\vspace{-0.5in}Experiences with the GTU grammar development
environment}
\author{Martin Volk \\
University of Zurich \\
Department of Computer Science \\
Computational Linguistics \\
Winterthurerstr. 190 \\
CH-8057 Zurich \\
{\tt volk@ifi.unizh.ch}
\And
Dirk Richarz \\
University of Koblenz-Landau \\
Institute of Computational Linguistics \\
Rheinau 1 \\
D-56075 Koblenz \\
{\tt richarz@informatik.uni-koblenz.de}
}

\begin{document}

\bibliographystyle{fullname}
\maketitle
\vspace{-0.5in}

\maketitle

\begin{abstract}
In this paper we describe our experiences with a tool for
the development and testing of natural language grammars called GTU
(German:
Grammatik-Testumgebumg; grammar test environment). GTU supports four
grammar
formalisms under a window-oriented user interface. Additionally, it
contains a
set of German test sentences covering various syntactic phenomena as
well as three types of German lexicons that can be attached to a grammar
via an integrated lexicon interface. What follows is a description of
the experiences we gained when we used GTU as a tutoring tool for
students and as an experimental tool for CL researchers. From these we
will derive the  features necessary for a future grammar workbench.
\end{abstract}

\section{Introduction}
GTU (German: Grammatik-Testumgebung; grammar test environment) was
developed
as a flexible and user-friendly tool for the development and testing of
grammars in various formats. Throughout the last 7 years it has been 
successfully used as a tutoring tool to supplement syntax courses in 
computational linguistics at the Universities of Koblenz and Zurich.  

GTU has been implemented in Arity Prolog under DOS and OS/2, and in
SICStus Prolog under UNIX. In this paper we will concentrate on the UNIX
version.  GTU in this version is a stand-alone system of about 4.5 MB
compiled Prolog code
(not counting the lexicons)\footnote{According to rearrangements 
of the operating system the actual memory requirements total
about 7 MB for both SUN OS 4.x and SUN OS 5.x.}. GTU interacts with 3
German lexicons:

\begin{enumerate}
\item a small hand-coded stem-lexicon whose vocabulary has been tailored
towards the test sentences (This lexicon also contains selectional
restrictions for all its nouns and adjectives.), 
\item GerTWOL \cite{Ling94}, a fast morphology analysis program, and
\item PLOD, a full-form lexicon that has been derived from the CELEX
lexical database  \cite{Cele95}.
\end{enumerate}

GTU supports grammars under four formalisms:
\begin{enumerate}

\item Definite Clause Grammar (DCG, \cite{Pere87}) augmented with
feature
      structures,
\item Immediate Dominance / Linear Precedence Grammar (ID/LP; a subset
of GPSG), 
\item Generalized Phrase Structure Grammar (GPSG, \cite{Gazd85a}),
\item Lexical Functional Grammar (LFG, \cite{Kapl82}).
 
\end{enumerate}

Additionally, GTU provides a first step towards semantic processing of
LFG
f-structures. Thus a grammar developer may specify the way the semantic
module computes logical expressions for an f-structure using semantic
rules.  In another module the selectional restrictions of the hand-coded
lexicon can be used to compute if (a reading of) a sentence is
semantically anomalous. This module can be switched on and off when
parsing a sentence.

GTU's features have been published before (see \cite{Jung94} or
\cite{Volk95b}). 
In this paper we concentrate on evaluating GTU's features, comparing
them to some other workbenches that we have access to (mostly GATE
\cite{Gaiz96} and the Xerox LFG workbench \cite{Kapl96}). From this we
derive recommendations for future grammar workbenches.

\section{GTU - its merits and its limits}

\subsection*{Grammar rule notation}

One of the primary goals in the GTU project was to support a grammar
rule notation that is as close as possible to the one used in the
linguistics literature. This has been a general guideline for every
formalism added to the GTU system. Let us give some examples. Typical
ID-rules in GTU are:

\begin{verbatim}
(1) S          -> NP[X],  
                  VP[X] | X = [kas=nom].
(2) NP[kas=K]  -> Det[kas=K, num=N], 
                  (AdjP[kas=K, num=N]), 
                  N[kas=K, num=N].
\end{verbatim}

Rule (1) says, that a constituent of type \verb+S+ consists of 
constituents of type \verb+NP+ and \verb+VP+. The feature
structures are given in square brackets. A capital letter in a feature
structure represents a variable. Identical variables within a rule stand
for
shared values. Hence, the feature structures for \verb+NP+ and
\verb+VP+ in rule (1) are declared to be identical. In addition the
feature
structure equation behind the vertical bar \verb+|+ specifies that 
\verb+X+ must be unified with the feature structure \verb+[kas=nom]+.
Rule (2) says that an \verb+NP+ consists of a \verb+Det+,  an optional
\verb+AdjP+ and an \verb+N+. It also says that the features \verb+kas+
and \verb+num+ are set to be identical across constituents while only
the feature \verb+kas+ is passed on to the \verb+NP+-node. 

There are further means for terminal symbols within a grammar and a
reserved word representing an empty constituent.

In our experience the grammar rule notation helps the students in
getting acquainted with the system. But students still need some time in
understanding the syntax. In particular they are sometimes misled by the
apparent similarity of GTU's ID-rules to Prolog DCG-rules. While in
Prolog constituent symbols are atoms and are usually written with lower
case letters, GTU requires upper case letters as is customary in the
linguistic literature. In addition students need a good understanding of
feature structure unification to be able to manipulate the grammatical
features within the grammar rules.

For writing grammar rules GTU has an integrated editor that facilitates
loading the grammar into GTU's database. A grammar thus becomes
immediately available for testing. Loading a grammar involves the
translation of a grammar rule into Prolog. This is done by various
grammar processors (one for each formalism). The grammar processors are
SLR parsers generated from metagrammars. There is one metagrammar for
each grammar formalism describing the format of all admissible grammar
rules and lexicon interface rules under this formalism.

Writing large grammars with GTU has sometimes lead to problems in
navigation through the grammar files. A grammar browser could be used to
alliviate these problems. The Xerox LFG-WB contains such a browser. It
consists of a clickable index of all rule heads (i.e.\ all defined
constituent symbols). Via this index the grammar developer can
comfortably access the rule definitions for a given constituent.

\subsection*{Static grammar checks}
For the different formalisms in GTU, different types of parsers are
produced. GPSG
grammars are processed by a bottom-up chart parser, DCG and LFG grammars
are
processed by top-down depth-first parsers. All parsers have specific
problems with some structural properties of a grammar, e.g.\ top-down
depth-first parsers may run into infinite loops if the grammar contains
(direct or indirect) left recursive rules.

Therefore GTU provides a static check for detecting left recursions.
This is
done by building up a graph structure. After processing all grammar
rules and inserting all possible edges into the graph, the grammar
contains a possible left recursion if this graph contains at least one
cycle. In a similar manner we can detect cycles within transitive LP
rules or within alias definitions. 

These checks have shown to be very helpful in uncovering structural
problems once a grammar has grown to more than two dozen rules. The
static checks in GTU have to be explicitly called by the grammar
developer. It would be better to perform these checks automatically any
time a grammar is loaded into the system.

A model for the employment of grammar checks is the workbench for affix
grammars introduced by \cite{Nede92}, which uses grammar checks in order
to report on inconsistencies (conflicts with well-formedness conditions
such as that every nonterminal should have a definition), properties
(such as LL(1)), and information on the overall grammar structure (such
as the is-called-by relation). 

\subsection*{Output in different granularities}
One of GTU's main features is the graphics display of parsing results.
All constituent structures can be displayed as parse trees. For
LFG-grammars GTU additionally outputs the f-structure. For DCG and GPSG
the parse tree is also displayed in an indented fashion with all
features used during the parsing process. Output can be directed into
one or multiple windows. The multiple window option facilitates the
comparison of the tree structures on screen. Parsing results can also be
saved into files in order to use them in documentations or for other
evaluation purposes. 

The automatic graphic display of parsing results is an important feature
for using GTU as a tutoring tool. For students this is the most striking
advantage over coding the grammar directly in a programming language.
The GTU display works with structures of arbitrary size. But a structure
that does not fit on the screen requires extensive scrolling. A zoom
option could remedy this problem.

Zooming into output structures is nicely integrated into the Xerox
LFG-WB. Every node in the parse tree output can be enlarged by a mouse
click to its complete feature structure. Every label on a chart edge
output can be displayed with its internal tree structure and with its
feature structure.

\subsection*{Automatic comparison of output structures}
When developing a grammar it often happens that the parser finds
multiple parses for a given sentence. Sometimes these parses differ only
by a single feature which may be hard to detect by a human. Automatic
comparison of the parses is needed. This can also be used to compare the
parses of a given sentence before and after a grammar modification. 

It is difficult to assess the effects of a grammar modification. Often
it is  necessary to rerun long series of tests. In these tests one wants
to save the parse structure(s) for a given test sentence if a certain
level of coverage and correctness has been reached. Should a
modification of the grammar become necessary, the newly computed parse
structure can be automatically compared to the saved structure. We have
included such a tool in GTU. 

The comparison tool works through three subsequent levels. First, it
checks whether the branching structures of two parse trees are
identical, then it compares the node names (the constituent symbols),
and finally it detects differences in the feature structures. The
procedure stops when it finds a difference and reports this to the user. 

Implementing such a comparison tool is not too difficult, but
integrating it into the testing module of a grammar workbench is a major
task, if this module supports different types of tests (single sentence
tests and series of tests; manual input and selections from the test
suite). At the same time one needs to ensure that the module's
functionality is transparent and its handling is easy. For example, what
should happen if a sentence had two readings before a grammar
modification and has three readings now? We decided to compare the first
two new structures with the saved structures and to inform the user that
there now is an additional reading. In our comparison tool series of
comparisons for multiple sentences can be run in the background. Their
results are  displayed in a table which informs about the numbers of
readings for every sentence. 

This comparison tool is considered very helpful, once the user
understands how to use it. It should be complemented with the option to
compare the output structures of two readings of the same input
sentence.

\subsection*{Tracing the parsing process}
Within GTU the parsing of natural language input can be traced on
various levels. It can be traced
\begin{itemize}
\item during the lexicon lookup process displaying the
morpho-syntactical
      information for every word,
\item during the evaluation of the lexicon interface rules displaying
the
      generated lexical rules for a given word,
\item during the application of the grammar or semantic rules.
\end{itemize}

For GPSG grammars GTU presents every edge produced by the bottom-up
chart
parser. For DCG and LFG grammars GTU shows ENTRY, EXIT, FAIL and REDO
ports for a predicate, as in a Prolog development environment. But
GTU does not provide options for selectively skipping the trace for a
particular category or for setting special interrupt points that allow
more goal-oriented tracing. Furthermore, the parser cannot be
interrupted by an abort option in trace mode. These problems lead to a
reluctance in using the trace options since most of the time too much
information is presented on the screen. Only elaborate trace options are
helpful in writing sizable grammars.

\subsection*{Lexicon interface}
The flexible lexicon interface is another of GTU's core elements. With
special lexicon interface rules that are part of every grammar formalism
the grammar developer can specify which lexicon information the grammar
needs and how this information should be structured and named.

For each word a lexicon provides information about the possible part of
speech   and morpho-syntactical information. Lexicon interface rules
determine how this information is passed to the grammar.

A lexicon interface rule contains a test criterion and a specification
and has
the following format:
\begin{quote}
\texttt{if\_in\_lex (\emph{test criterion}) then\_in\_gram
(\emph{specification}) }.
\end{quote}

The test criterion is a list of feature-value pairs to be checked
against a word's lexical information. Additionally, constraints are
allowed that check if some feature has a value for the given word. For
example, the test  
\begin{verbatim}
(pos=verb, !tense, ~reflexive)
\end{verbatim}
will only succeed for irreflexive finite verbs\footnote{'!feature' means
that the feature must have some value, while '$\sim$feature' prohibits
any value on the feature.}.

While it is necessary that the test contains only features available in
the lexicon, the specification part may add new information to the
information found in the lexicon. For example, the specification
\begin{verbatim}
case = #kasus, number = #numerus, person = 3
\end{verbatim}
assigns the value of the feature \texttt{kasus} found in the lexicon
(which is indicated by \verb'#') to a feature named \texttt{case} (and
the like for number). Additionally, a new feature \texttt{person} is
added with the value \texttt{3}. In this way every noun may get a
specification for the \texttt{person} feature.

The specification part defines how lexicon information shall be mapped
to a
syntactic category in case the test criterion is met. While the format
of the test criterion is the same for all formalisms, the format of the
specification has been adjusted to the format of every grammar
formalism. In this way the definition of lexical entries can be adapted
to a grammar formalism while reusing the lexical resources. 

Writing lexicon interface rules requires a good understanding of the
underlying lexicon. And sometimes it is difficult to see if a problem
with lexical features stems from the lexicon or is introduced by the
interface rules. But overall this  lexicon interface has been
successful. With its simple format of rules with conditions and
constraints it can serve as a model for interfacing other modules to a
grammar workbench.

\subsection*{Test suite administration}
GTU contains a test suite with about 300 sentences annotated with their
syntactic properties. We have experimented with two representations of
the test suite \cite{Volk95}. One representation had every sentence
assigned to a phenomenon class and every class in a separate file. Each
sentence class can be loaded into GTU and can be separately tested. In a
second representation the sentences were organized as leaves of a
hierarchical tree of syntactic phenomena. That is, a phenomenon like
'verb group syntax' was subdivided into 'simple verb groups', 'complex
verb groups', and 'verb groups with separated prefixes'. The sentences
were attached to the phenomena they represented. In this representation
the grammar developer can select a phenomenon resulting in the display
of the set of subsumed sentences. If multiple phenomena are selected 
the intersection of the sets is displayed. 

It turned out that the latter representation was hardly used by our
students. It seems that grammar writing itself is such a complex process
that a user does not want to bother with the complexities of navigating
through a phenomena tree. The other, simple representation of sentence
classes in files is often used and much appreciated. It is more
transparent, easier to select from, and easier to modify (i.e.\ it is
easier to add new test sentences).

Few other grammar workbenches include an elaborate test module and only
PAGE \cite{Oepe97} comprises a test suite which is integrated similarly
to GTU. PAGE's test suite, however, is more comprehensive than GTU's
since it is based on the TSNLP (Test Suites for Natural Language
Processing) database. TSNLP provides more than 4000 test items for
English, French and German each. We are not aware of any reports of this
test suite's usability and acceptability in PAGE.

\subsection*{Output of recognized fragments in case of ungrammaticality}
In case a parser cannot process the complete natural language input, it
is mandatory that the grammar developer gets feedback about the
processed fragments. GTU presents the largest recognized fragments. That
is, starting from the beginning of the sentence it takes the longest
fragment, from the end of this fragment it again takes the longest
fragment and so on. If there is more than one fragment of the same
length, only the last one parsed is shown. The fragments are retrieved
from the chart (GPSG) or from a well-formed substring table (DCG, LFG).
Obviously, such a display is sometimes misleading since the selection is
not based on linguistic criteria.

As an alternative we have experimented with displaying the shortest
paths through the chart (i.e.\ the paths from the beginning to the end
of the input with the least number of edges). In many cases such a path
is a candidate close to a parsing solution. In general, it fares better
than the longest fragments but again it suffers from a lack of
linguistic insight. 

Yet another way is to pick certain combinations of constituents
according to predefined patterns. It is conceivable that the grammar
developer specifies an expected structure for a given sentence and that
the system reports on the parts it has found. Or the display system may
use the grammar rules for selecting the most promising chart entries.
Displaying the complete chart, as done in the Xerox LFG-WB, will help
only for small grammars. For any sizable grammar this kind of display
will overwhelm the user with hundreds of edges.

Selecting and displaying chart fragments is an interesting field where
more research is urgently needed, especially with respect to treating
the results of parsing incomplete or ill-formed input.

\subsection*{Lexicon extension module}
When writing grammars for real natural language sentences, every
developer will soon encounter words that are not in the lexicon,
whatever size it has. Since GTU was meant as a tutoring tool it contains
only static lexicons. In fact, its first lexicon was tailored towards
the vocabulary of the test suite. GTU does not provide an extension
module for any of the attached lexical resources. The grammar developer
has to use the information as is. Adding new features can only be done
by inserting them in lexicon interface rules or grammar rules. Words can
be added as terminal symbols in the grammar.

This is not a satisfactory solution. It is not only that one wants to
add new words to the lexicon but also that lexicon entries need to be
corrected and that   new readings of a word need to be entered. In that
respect using GerTWOL is a drawback, since it is a closed system which
cannot be modified. (Though its developers are planning on extending it
with a module to allow adding 
words.\footnote{Personal communication with Ari Majorin of Lingsoft,
Helsinki, in December 1996.}) The other lexicons within GTU could in
principle be modified, and they urgently need a user interface to
support this. This is especially important for the PLOD-lexicon derived
from the CELEX lexical database, which contains many errors and
omissions. 

Models for lexicon extension modules can be found in the latest
generation of commercial machine translation systems such as IBM's
Personal Translator or Langenscheidts T1. Lexicon extension in these
systems is made easy by menus asking only for part of speech and little
inflectional information. The entry word is then classified and all
inflectional forms are made available. 

Of course in a multi-user system these modifications need to be
organized with access restrictions. Every developer should be able to
have his own sublexicon where lexicon definitions of any basic lexicon
can be superseded. But only a designated user should be allowed to
modify the basic lexicon according to suggestions sent to him by the
grammar developers.

\subsection*{Combination of lexical resources}

GTU currently does not support the combination of lexical resources.
Every lexical item is taken from the one lexicon selected by the user.
Missing features cannot be complemented by combining lexicons. This is a
critical aspect because none of the lexicons contains every information
necessary. While GerTWOL analyzes a surprising variety of words and
returns  morphological information with high precision, it does not
provide any syntactical information. In particular it does not provide 
a verb's subcategorization. This information can be found in the
PLOD/CELEX lexicon to some degree. For example, the grammar developer
can find out whether a verb requires a prepositional object, but he
cannot find out which preposition the phrase has to start
with.\footnote{The next version of CELEX will contain such prepositional
requirements. (Personal communication with CELEX manager Richard
Piepenbrock in April 1997)}

\subsection*{Clear modularization}
The development of a large grammar  - like a large software system - 
makes it necessary to split the work into modules. GTU supports such
modularisation into files that can be loaded and tested independently.
But GTU only recommends to divide a grammar into modules, it does not
enforce modularisation. For a consistent development of large grammars,
especially if distributed over a group of people, we believe that a
grammar workbench should support more engineering aspects we know from
software development environments such as a   module 
concept with clear information hiding, visualisation of call graphs on
various levels, or summarisation of selected rule properties.

\subsection*{General remarks on GTU}
GTU focuses on grammar writing. It does not include any means to
influence parsing efficiency. But parsing efficiency is another
important aspect of learning to deal with grammars and to write NLP
systems. It would therefore be desirable to have a system with
parameterizable parsers. On the other hand this might result in an
unmanageable degree of complexity for the user and - like with the
alternative test suite - we will end up with a nice feature that nobody
wants to use. 

The GTU system has been implemented with great care. Over time more than
a dozen programmers have contributed modules to the overall system. The
robust integration of these modules was possible since the core
programmers did not change. They had documented their code in an
exemplary way. Still, the problem of interfacing new modules has
worsened. A more modular approach seems desirable for building large
workbenches.

\section{Different grammar development environments}

In order to position GTU within the context of grammar development
environments, let us classify them according to their purpose. 

\begin{description}
\item[Tutoring environments] are designed for learning to write
grammars. They must be robust and easy to use (including an intuitive
format for grammar rules and an intuitive user interface). The grammar
developer should be able to focus on grammar writing. Lexicon and test
suite should be hidden. Tutoring environments therefore should contain a
sizable lexicon and a test suite with a clear organisation. They should
provide for easy access to and intuitive display of intermediate and
final parsing results. They need not bother with efficiency
considerations of processing a natural language input. GTU is an example
of such a system.

\item[Experimentation environments] are designed for professional
experimen\-tation and demonstration. They must also be robust but they
may require advanced engineering and linguistic skills. They should
provide for checking the parsing results. They must support the grammars
and parsers to be used outside the development system. We think that
Alvey-GDE \cite{Carr91} and Pleuk \cite{Cald93} are good examples of
such environments. They allow the tuning of the parser (Alvey) and even
redefining the grammar formalism (Pleuk). 
The Xerox LFG-WB is partly a tutoring environment (especially with its
grammar index and zoom-in displays) and partly an experimentation
environment since it lacks a test suite and a lexicon. 

Note that the systems also differ in the number of grammar formalisms
they support. The Alvey-GDE (for GPSG) and the Xerox LFG-WB work only
for one designated formalism. GTU has built-in processors for three
formalisms, and Pleuk supports whatever formalism one defines. 

\item[NLP environments] are designed as platforms for the development of
multi-module NLP systems. Rather than being a closed system they provide
a shell for combining multiple linguistic modules such as tokenizers,
taggers, morphology analyzers, parsers (with grammars) and so on. A
grammar workbench is a tool to develop such a module. All the modules
can be tailored and tuned to the specific needs of the overall system.
We consider ALEP \cite{Simp94} and GATE \cite{Gaiz96} to be examples of
such environments. Although it seems logical and desirable that NLP
environments should provide for the delivery of stand-alone systems this
aspect has been neglected so far. In particular we suspect that the
interface format, as required e.g.\ between GATE modules, will have
negative effects on the processing efficiency of the complete
system.\footnote{GATE requires modules to communicate via a so called
CREOLE interface, which is a layer wrapped around an existing module.} 
\end{description}

GTU was designed as a tutorial system for grammar development. Over time
it has grown into a system that supports most functions of
experimentation environments. Its main limitations are its closed
architecture and the inability to use the grammars outside the system.
Many of its modules can be employed by an NLP environment. GTU's most
successful modules are its flexible lexicon interface, the tight
integration of the test suite and the module for comparison of output
structures.

An NLP environment should be an open platform rather than a closed
workbench, as is the core concept of ALEP and GATE. This is needed to
allow special treatment for special linguistic problems. For instance,
the treatment of separable prefix verbs in German is so specific that it
could be tackled by a preprocessor before parsing starts. Only after the
separated prefix and the main verb have been recompounded the verb's
subcategorization can be determined.

Another specific problem of German is the resolution of elliptical
coordinated compounds (e.g.\ {\em In- und Ausland} standing for {\em
Inland und Ausland}). If such ellipses are filled in before parsing
starts such a coordination does not need special grammar rules. Other
peculiarities such as date, time, currency, distance expressions will
also need special modules. In this way only 
the processing of the core syntactic phenomena is left to the parser.

An NLP environment should allow parametrisation of parsers or multiple
parsers of different processing strategies (e.g.\ a combination of
symbolic and statistic parsing) and processing depths (e.g.\ shallow
parsing if no complete parse can be found).

\section{Conclusions}
Tools like GTU are well suited for learning to develop grammars, for
experimenting with grammar formalisms, and for demonstrating the work of
computational linguistics. The use of GTU as an educational tool in
computational linguistics courses has been very successful. In a recent
project GTU's flexibility is being challenged in a joint project with
the Institute of Germanic Language at the University of Koblenz. In this
project we examine
the use of GTU for the benefit of courses in German as a foreign
language.

For the development of large grammars in combination with large
linguistic resources and for processing them efficiently, GTU is less
suited. We are now convinced that we need an open platform that provides
a framework for combining modules for such a task. For this it is
necessary to develop interface standards for different types of modules
(taggers, grammars, lexicons, test suites etc.).

Finally, we should keep in mind that a computational environment for
grammar development offers help in engineering NLP modules for
well-understood phenomena. The real hard problems in NLP (most
importantly the resolution of ambiguity) need to be solved by bringing
to bear the right information at the right time. But this is of yet a
complex area with many questions that have not received a theoretical
answer let alone an engineering solution.

\section{Acknowledgements}
We would like to thank Diego Moll\`a Aliod and Gerold Schneider for
providing background information on some grammar workbenches.


\begin{thebibliography}{}

\bibitem[\protect\citename{Baayen, Piepenbrock, and van
Rijn}1995]{Cele95}
Baayen, R.~H., R.~Piepenbrock, and H.~van Rijn.
\newblock 1995.
\newblock The {CELEX} lexical database ({CD-ROM}).
\newblock Linguistic Data Consortium, University of Pennsylvania.

\bibitem[\protect\citename{Calder and Humphreys}1993]{Cald93}
Calder, J. and K.~Humphreys.
\newblock 1993.
\newblock Pleuk overview.
\newblock Technical report, University of Edinburgh. Centre for
Cognitive
  Science.

\bibitem[\protect\citename{Carroll, Briscoe, and Grover}1991]{Carr91}
Carroll, John, Ted Briscoe, and Claire Grover.
\newblock 1991.
\newblock A development environment for large natural language grammars.
\newblock Technical report, University of Cambridge Computer Laboratory.

\bibitem[\protect\citename{Gaizauskas \bgroup et al.\egroup
}1996]{Gaiz96}
Gaizauskas, R., H.~Cunningham, Y.~Wilks, P.~Rodgers, and K.~Humphreys.
\newblock 1996.
\newblock {GATE}: an environment to support research and development in
natural
  language engineering.
\newblock In {\em Proc. of the 8th IEEE Conf. on tools with AI
(ICTAI-96)}.

\bibitem[\protect\citename{Gazdar \bgroup et al.\egroup }1985]{Gazd85a}
Gazdar, Gerald, Ewan Klein, Geoffrey Pullum, and Ivan Sag.
\newblock 1985.
\newblock {\em Generalized phrase structure grammar}.
\newblock Harvard University Press, Cambridge,MA.

\bibitem[\protect\citename{Jung, Richarz, and Volk}1994]{Jung94}
Jung, Michael, Dirk Richarz, and Martin Volk.
\newblock 1994.
\newblock {GTU} - {E}ine {G}rammatik-{T}estumgebung.
\newblock In {\em Proceedings of KONVENS-94}, pages 427--430, Wien.

\bibitem[\protect\citename{Kaplan and Maxwell}1996]{Kapl96}
Kaplan, R.M. and J.T.~Maxwell III, 1996.
\newblock {\em LFG Grammar Writer's Workbench (Version 3.1)}.
\newblock Xerox Corporation.

\bibitem[\protect\citename{Kaplan and Bresnan}1982]{Kapl82}
Kaplan, Ronald and Joan Bresnan.
\newblock 1982.
\newblock Lexical-functional grammar. {A} formal system for grammatical
  representation.
\newblock In Joan Bresnan, editor, {\em The Mental Representation of
  Grammatical Relations}. MIT Press, Cambridge,MA.

\bibitem[\protect\citename{Nederhof \bgroup et al.\egroup }1992]{Nede92}
Nederhof, M.J., C.H.A. Koster, C.~Dekkers, and A.~van Zwol.
\newblock 1992.
\newblock The grammar workbench: A first step towards lingware
engineering.
\newblock In W.~ter Stal, A.~Nijholt, and R.~op~den Akker, editors, {\em
  Proceedings of Second Twente Workshop on Language Technology}, pages
  103--115, Twente, NL.

\bibitem[\protect\citename{Oepen}1997]{Oepe97}
Oepen, Stephan.
\newblock 1997.
\newblock {PAGE. Platform for Advanced Grammar Engineering}.
\newblock WWW page (http://cl-www.dfki.uni-sb.de/cl/systems /page),
April.

\bibitem[\protect\citename{Oy}1994]{Ling94}
Oy, Lingsoft.
\newblock 1994.
\newblock {Gertwol. Questionnaire for Morpholympics 1994}.
\newblock {\em LDV-Forum}, 11(1):17--29.

\bibitem[\protect\citename{Pereira and Shieber}1987]{Pere87}
Pereira, Fernando~C.N. and Stuart~M. Shieber.
\newblock 1987.
\newblock {\em Prolog and Natural-Language Analysis}, volume~10 of {\em
CSLI
  Lecture Notes}.
\newblock University of Chicago Press, Stanford.

\bibitem[\protect\citename{Simpkins}1994]{Simp94}
Simpkins, N.K.
\newblock 1994.
\newblock {An open architecture for language engineering. The Advanced
Language
  Engineering Platform (ALEP)}.
\newblock In {\em Proceedings of Language Engineering Convention,
Paris}.
  European Network in Language and Speech, Centre for Cognitive Science,
  Edinburgh, pages 129--136.

\bibitem[\protect\citename{Volk, Jung, and Richarz}1995]{Volk95b}
Volk, M., M.~Jung, and D.~Richarz.
\newblock 1995.
\newblock {GTU - A workbench for the development of natural language
grammars}.
\newblock In {\em Proc. of the Conference on Practical Applications of
Prolog},
  pages 637--660, Paris.

\bibitem[\protect\citename{Volk}1995]{Volk95}
Volk, Martin.
\newblock 1995.
\newblock {\em Einsatz einer {T}estsatzsammlung im {G}rammar
{E}ngineering},
  volume~30 of {\em Sprache und Information}.
\newblock Niemeyer Verlag, T\"ubingen.

\end{thebibliography}
\end{document}